\documentclass[10pt,twocolumn,letterpaper]{article}

\usepackage{iccv}              % To produce the CAMERA-READY version
\usepackage{layout}
\usepackage{siunitx}
\DeclareMathOperator*{\argmax}{arg\,max}

\definecolor{iccvblue}{rgb}{0.21,0.49,0.74}
\usepackage[pagebackref,breaklinks,colorlinks,allcolors=iccvblue]{hyperref}

\def\paperID{3321} % *** Enter the Paper ID here
\def\confName{ICCV}
\def\confYear{2025}

\title{Visual Surface Wave Elastography:\\Revealing Subsurface Physical Properties via Visible Surface Waves}

\author{Alexander C. Ogren$^{*}$ \quad Berthy T. Feng$^{*}$ \quad Jihoon Ahn \quad Katherine L. Bouman \quad Chiara Daraio \\
{California Institute of Technology}
}
\begin{document}
% \layout
\maketitle
\begin{abstract}

% \vspace{-1mm}
Wave propagation on the surface of a material contains information about physical properties beneath its surface.
We propose a method for inferring the thickness and stiffness of a structure from just a video of waves on its surface.
Our method works by extracting a dispersion relation from the video and then solving a physics-based optimization problem to find the best-fitting thickness and stiffness parameters.
We validate our method on both simulated and real data, in both cases showing strong agreement with ground-truth measurements.
Our technique provides a proof-of-concept for at-home health monitoring of medically-informative tissue properties, and it is further applicable to fields such as human-computer interaction.

\end{abstract}
{\let\thefootnote\relax\footnote{{*equal contribution}}}
\section{Introduction}
% Outline
% 1. Why sense material properties passively
% 2. Material characterization for at-home health monitoring
% 3. Key idea of proposed approach
% 4. Proposed approach
% 5. Previous work
% 6. Summary of contributions

How waves propagate on the surface of a medium reveals information about its subsurface properties.
%v1
% In the ocean, the waves we see are only the tip of the iceberg - the dynamics of these waves extend well below the surface, and by watching how they change and crash as they get closer to shore, you can get a sense for the gradual upward slope of the ocean floor below.
%v2
For example, by watching how ocean waves evolve and break as they near the shore, one can infer the rises and dips of the seafloor below \cite{webb2021introduction}.
%v0
% Imagine throwing a pebble into a pond, which sends ripples through the water.
% Even though you can only see these waves on the surface, they contain clues to the depth of the pond and the murkiness of its water.
% \alex{Maybe I could stitch in a piece here about how ocean waves change and crash as they become closer to shore? It would be good foreshadowing for "thickness is important"}
% For example, you might vaguely sense that the pond is very deep if the wavelengths are large.
Waves can also be observed on the surfaces of biological systems.
Imagine applying a massage gun to your calf.
% The ripples on your skin convey information about the biological material below that dictates their dynamics.
The ripples on your skin convey information about the underlying layers of fat, muscle, and bone.
% the thickness and stiffness of each layer of fat or muscle.
In fact there is a well-defined relationship linking the thickness and stiffness of each layer to the wave propagation behavior.
We propose \textit{Visual Surface Wave Elastography} (VSWE), a physics-based method to estimate the thickness and stiffness of a medium from a video of waves on its surface.

\begin{figure}
    \centering
    \includegraphics[width=\linewidth]{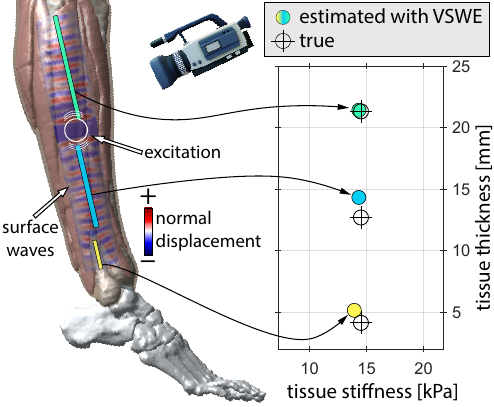}
    \caption{Estimating subsurface tissue properties of a human leg with VSWE. From a video of surface wave motion, VSWE estimates the thickness and stiffness of the soft tissue layer.
    In this example, we are able to recover the three different thicknesses of the three regions highlighted on the leg. We also recover the stiffness of the leg tissue, which does not change across the leg.
    \vspace{-0.1in}
    }
    \label{fig:teaser}
\end{figure}

We are primarily motivated by the task of biological tissue characterization, which has broad applications including at-home health monitoring.
For example, tumors \cite{hoyt2008tissue,kwonComparisonCancerCell2020}, musculoskeletal degeneration \cite{varadyOsteoarthritisYearReview2016,lacourpailleNoninvasiveAssessmentMuscle2015,virgilioMultiscaleModelsSkeletal2015}, and liver disease \cite{georgesIncreasedStiffnessRat2007,sunBiomechanicsFunctionalityHepatocytes2014a} often lead to changes in tissue thickness or stiffness.
% For example, tumors often stiffen the tissues they inhabit \cite{hoyt2008tissue,kwonComparisonCancerCell2020}; degenerative diseases like osteoarthritis \cite{varadyOsteoarthritisYearReview2016} and muscular dystrophy \cite{lacourpailleNoninvasiveAssessmentMuscle2015,virgilioMultiscaleModelsSkeletal2015} lead to the softening of cartilage and muscle tissue, respectively; and liver diseases such as fibrosis and cirrhosis are characterized by a progressive stiffening of the liver \cite{georgesIncreasedStiffnessRat2007,sunBiomechanicsFunctionalityHepatocytes2014a}.
However, existing techniques for elastography (i.e., measuring tissue stiffness) require specialized ultrasound \cite{wongTransientElastography2010,taljanovicShearWaveElastographyBasic2017} and sometimes magnetic resonance \cite{mariappanMagneticResonanceElastography2010} devices, along with medical experts to operate the equipment \cite{bellEconomicsMRITechnology1996,bureau2016economics}.
We show that it is possible to obtain coarse estimates of thickness and stiffness given surface waves measured with just a video camera.
% \alex{Could I also insert a short piece here about using this for bio-sensing? Could mention that it could be used for detecting changes in stiffness over a short time period}
% Furthermore, our technique has applications in fields besides medicine.
% For example, visually detecting real-time changes in tissue stiffness may be of interest for the future of human-computer interaction.
% While many of today's common computing devices leverage input from sensors directly contacting the skin (keyboard, mouse, even EMG sensors), some less common forward-looking computing paradigms such as smartglasses and augmented-reality headsets are positioning to rely more and more on visual data to reduce the need for worn sensors.
% Visually inferring that a certain muscle is flexed (via sensing a change in stiffness from its relaxed state) may unlock new predictive information in gesture recognition, paving new ways to interact with worn computers. \alex{hmm I kind of want to compactify this XR section}
Another application is human computer interaction (HCI), where visually inferring subsurface changes in muscle stiffness may unlock new modes of gesture recognition.
In biomechanics, many regions of the body are often modeled as a series of layers of biological tissue, such as skin, muscle, or bone \cite{layermodelexample,geldof2022layer}.
% for which visually inferring that a certain muscle is flexed (via sensing a change in stiffness from its relaxed state) may unlock new predictive information in gesture recognition.
% COMMENTED OUT SEISMOLOGY APPLICATION BELOW
% Our work is also relevant to geophysics and seismology, where measurements of seismic waves help scientists recover underground geological features, as well as structural health monitoring and non-destructive testing, which aim to non-destructively infer geometric and mechanical object properties.
Similarly, in this work, we model the medium as a layer of soft tissue with unknown thickness and stiffness atop a layer of bone.
% This model is accurate for regions of layers with uniform thickness, such as the upper leg or upper arm.
% We can target more complex anatomical structures by dividing it into sub-regions in which the layered model holds (e.g., by dividing the lower leg into the calf and the Achilles and applying our method separately to each region).
% Adopting a simplified geometric model allows us to develop a computational solution to an otherwise extremely challenging inverse problem of inferring both geometry and stiffness throughout an object.
We leverage the mathematical relationship between the geometric and mechanical properties of the soft layer and the wave propagation behavior of the medium.

The wave propagation behavior of a material is compactly described by a mathematical object known as a \textit{dispersion relation}.
Under some common biomechanical assumptions in our layer model (discussed further in \cref{sec:dispersion_relations}), the thickness and stiffness of the soft layer fully determine the dispersion relation.
% Conveniently, the thickness and material behavior of each layer fully determine the dispersion relation.
% Under some common biomechanical assumptions, the material behavior can be fully described by its stiffness.
The main idea of our method is to find the thickness and stiffness values that lead to a dispersion relation that matches the dispersion relation extracted from surface waves observed in the video.

% Previous work
% Our work has several advantages over previous non-destructive testing methods.
% First of all, we only require a video camera instead of a specialized sensor like a laser vibrometer \cite{durst1976principles} \TODO{anything else?} to measure surface displacements. 
% Although techniques exist to measure small motions in videos \cite{wadhwa2013phase,davis2015visual}, these ``visual vibrometry'' methods use vibrational modes \cite{feng2022visual,visvib2017pami,davis2015image} rather than surface wave propagation.
% We propose a technique for leveraging surface waves, which are more appropriate to use in certain scenarios.
% \TODO{really need to figure out when surface waves are more appropriate}
% \alex{maybe we can argue that modal information generally needs to be global (even though we didn't observe all sides, we observe the whole object in a way) while waves can be observed locally at different points}

% Summary of contributions
In this paper, we first discuss related work on video-based and wave-based material characterization and provide background on wave mechanics.
We then describe our method, which consists of taking a video of surface waves, extracting the dispersion relation from the video, and then solving an optimization problem to find the best-fitting thickness and stiffness parameters.
We validate our method on real and simulated data, including real videos of gelatin-based phantoms that mimic biological tissue and realistic simulations of a human leg.

% Our work demonstrates the promise of inferring critical material properties from simply a video of surface wave motion.
\section{Related work}
\subsection{Video-based material characterization}
Videos encode rich information about the environment and physical objects.
For example, previous work has leveraged surface vibrations \cite{Davis2014VisualMic} or sound \cite{feng2023self,chen2023sound} in video to recover unknown information.
For material characterization tasks, the way an object moves in a video provides useful information about its physical properties.
% Some methods estimate material parameters from video through data-driven techniques, without the use of principled physics-based models \cite{bouman2013estimating}.
% % Some methods estimate parameters of specific object classes, such as fabrics \cite{bhat2003estimating, miguel2012data, bouman2013estimating} and trees \cite{wang2017botanical}, without analyzing dynamics in image-space.
% % Videos have been used to estimate material parameters for some specific object classes, such as fabrics \cite{bhat2003estimating, miguel2012data, bouman2013estimating} and trees \cite{wang2017botanical}.
% Other methods assimilate physically-grounded hypotheses to the observed data \cite{davis2015visual, visvib2017pami, miguel2012data, bhat2003estimating}
% Another category of methods involves extracting surface displacements in image-space (through phase-based motion processing \cite{wadhwa2013phase} or optical flow \cite{beauchemin1995computation}) and then using these measurements to constrain the estimation of physical parameters.
% For example, Visual Vibrometry \cite{davis2015visual, visvib2017pami} uses the image-space motion spectrum to estimate homogenized stiffness and damping parameters of fabrics and rods.
% Visual Vibration Tomography \cite{feng2022visual} estimates spatially-varying stiffness and density throughout an object with known geometry by solving a physics-based optimization problem involving 3D vibrational modes of the object.
Videos have been used to estimate material parameters of specific object classes, such as fabrics \cite{bhat2003estimating, miguel2012data, bouman2013estimating}, rods \cite{davis2015visual, visvib2017pami}, and trees \cite{wang2017botanical}, whose dynamics are entirely visible.
% Relevant to our work, one type of approach \cite{bouman2013estimating,davis2015visual, visvib2017pami} involves computing surface displacements in image-space (through phase-based motion processing \cite{wadhwa2013phase} or optical flow \cite{beauchemin1995computation}) and then using these measurements to constrain the estimation of physical parameters.
In contrast, Visual Vibration Tomography \cite{feng2022visual} infers \textit{subsurface} material properties.
Specifically, it recovers the spatially-varying stiffness and density throughout a 3D object with known geometry by analyzing its global vibrational modes.

Surface waves have been under-utilized for video-based material characterization, even though they can also be observed in video and contain useful information about underlying physical properties.
Our work shows how to leverage surface waves in video to estimate geometric and mechanical properties.
A surface-wave-based approach is beneficial when one wishes to analyze local structure without having to model the global structure, which may be complicated.
In the case of biological tissue characterization, for example, it would be impractical to model the 3D vibrational modes of the entire human anatomy in order to estimate local tissue properties.
Our surface-wave-based approach circumvents the need to model a complex geometry and solve for its global modes by targeting local regions where we can analyze wave modes of a simpler geometry.

\subsection{Wave-based material characterization}
In general, wave-based imaging relies on an understanding of how physical characteristics of interest affect wave propagation.
For example, magnetic resonance imaging \cite{hashemi2012mri} uses knowledge of how radio-frequency waves are absorbed and re-emitted by different types of tissues in the body.
Ultrasound techniques, which are pervasive in non-destructive testing \cite{jodhani2023ultrasonic}, medical imaging \cite{avola2021ultrasound}, and wearable technology \cite{huang2023emerging}, use high-frequency mechanical bulk waves to characterize tissue and locate features.
Among tissue characterizaton techniques, transient elastography \cite{wongTransientElastography2010}, shear wave elastography \cite{taljanovicShearWaveElastographyBasic2017}, and magnetic resonance elastography \cite{mariappanMagneticResonanceElastography2010} leverage knowledge of bulk wave physics to estimate the elasticity of tissues and organs.
However, such methods often require not only high-end, expensive equipment, but also trained medical specialists \cite{bellEconomicsMRITechnology1996,bureau2016economics}, making regular screening infeasible.
In contrast, our work leverages surface waves, which are generally less expensive to observe than their bulk counterparts.

Although very different fields from our application of interest, geophysics and seismic imaging provide insights into the possibility of using surface waves to infer subsurface features \cite{parkMASWGeotechnicalSite2013,shapiroHighResolutionSurfaceWaveTomography2005,gucunskiNumericalSimulationSASW1992,linDispersionAnalysisSurface2017,nazarianEvaluationModuliThicknesses1983}.
% While some researchers have extended these ideas to inexpensive tissue characterization \cite{bialasViscoelasticCharacterizationThin2011}, they often rely on relatively small arrays of sensors mounted in specific locations rather than leveraging full field data that can be acquired e.g. by video.
% While some researchers have extended these ideas to inexpensive tissue characterization \cite{bialasViscoelasticCharacterizationThin2011}, they rely on relatively small arrays of sensors mounted in specific locations on the skin rather than leveraging full field data that can be acquired e.g. by video.
% Previous work \cite{bialasViscoelasticCharacterizationThin2011} extended these surface-wave measurements taken by a small set of sensors placed on the skin.
% However, surface-wave-based techniques for biological tissue characterization.
Inspired by the success of surface-wave methods in seismology, we propose to harness surface waves observed in video to infer subsurface tissue properties.
Whereas previous work \cite{bialasViscoelasticCharacterizationThin2011} suggested an approach using sensors sparsely placed on the skin, we leverage dense visual data.
Vision-based tissue characterization would enable the next generation of health monitoring systems that take advantage of the ubiquity of visual sensors.
\section{Background}

\subsection{Dispersion relations}
\label{sec:dispersion_relations}
% TRY 5
Just as sound waves can be expressed as the combination of simple harmonic modes, waves traveling through any medium can be expressed as the combination of wave modes of different spatial and temporal frequencies.
A \textit{dispersion relation} compactly describes wave propagation by defining all the possible wave modes of a medium.
Specifically, it defines the \textit{wavevector} and \textit{frequency} of each possible wave mode, where the wavevector and frequency indicate the spatial and temporal rates of oscillation, respectively.

Mathematically, the dispersion relation can be determined by solving the harmonic elastic wave equation (a PDE) subject to phase-shifted periodic boundary conditions.
Let $\mathbf{x}=[x,y,z]^\intercal$ denote the spatial location and $\mathbf{u}(\mathbf{x})=[u(\mathbf{x}), v(\mathbf{x}), w(\mathbf{x})]^\intercal$ denote the displacement at $\mathbf{x}$.
Assuming an isotropic linear-elastic material, the harmonic elastic wave equation can be written as
\begin{equation}
    \label{eq:harmonic_elastic_wave_equation}
    - \omega^2 \mathcal{M}(\mathbf{u})(\mathbf{x}) = \mathcal{K}(\mathbf{u})(\mathbf{x})\qquad\forall\mathbf{x}\in\Omega,
\end{equation}
where $\mathcal{M}$ and $\mathcal{K}$ denote linear operators representing the mass and stiffness of the medium. 
The mass operator $\mathcal{M}$ depends on the density field $\rho(\mathbf{x})$.
The stiffness operator $\mathcal{K}$ depends on the elastic modulus $E(\mathbf{x})$ and Poisson's ratio $\nu(\mathbf{x})$.
The eigenvector solution $\mathbf{u}(\mathbf{x})$ describes the shape of the wave mode, and the eigenfrequency $\omega$ defines the temporal frequency of the wave.
Note that the solutions also depend on the domain $\Omega$, which in our case is parameterized by the thickness and length of the tissue layer.

When solving \cref{eq:harmonic_elastic_wave_equation}, boundary conditions are necessary to impose real-life assumptions of the physical system. 
In our setting, we assume that $\Omega$ is a finite subregion of the full wave medium (e.g., a section of a leg).
Furthermore, to simplify our analysis, we target waves traveling in one direction (i.e., along the $x$ direction).
This leads us to apply 1D periodic boundary conditions on the domain $x\in[0,a]$ to impose the assumption that the wave medium continues past the boundary of $\Omega$.

Specifically, we apply the Bloch-Floquet \cite{floquet1883equations,bloch1929quantenmechanik} (a.k.a. phase-shifted) periodic boundary conditions, which define a boundary condition for every wavenumber $\gamma\in[0,\pi/a]$.
We note that in general $\gamma$ is a wavevector, but in the case of 1D wave analysis, it is a wavenumber that indicates the number of wavelengths per unit length.
For a given $\gamma$, the boundary condition is defined as
\begin{equation}
    \label{eq:bloch_floquet_periodic_boundary_condition}
    \mathbf{u}(x=a,y,z) = \mathbf{u}(x=0,y,z) e^{\mathbf{i} \gamma a}.
\end{equation}
In words, the solution at one end ($x=0$) only differs from the solution at the other end ($x=a$) by the phase shift $\gamma a$.

To compute a dispersion relation, we solve \cref{eq:harmonic_elastic_wave_equation} subject to \cref{eq:bloch_floquet_periodic_boundary_condition} for $\gamma\in[0,\pi/a]$. For each $\gamma$, we solve a generalized eigenvalue problem and obtain multiple $(\omega, \mathbf{u}(\mathbf{x}))$ solution pairs. The dispersion relation is exactly the set of all $\omega$ solutions for every $\gamma$.

\subsection{Assumptions for tissue characterization}
For biological tissue characterization, we can model the medium as a layer of soft tissue atop a hard bone layer, where the top layer has uniform thickness $T$.
We note that throughout this work, we refer to the elastic modulus of the tissue as its stiffness.
We make several assumptions to reduce the complexity of the problem:
\begin{enumerate}
    \item The soft layer has a uniform stiffness $E(\mathbf{x})=E$.
    \item The density and Poisson's ratio of the soft tissue are known. We set $\rho=1$\unit{\gram\per\cubic\centi\metre} \cite{SZEMEL2012461} and $\nu=0.45$ \cite{javanmardi2021quantifying}.
    \item The bone is much stiffer than the soft tissue.
    This means neither the thickness nor the exact stiffness of the bone layer matters, and we can model it as motionless.
\end{enumerate}
These assumptions are common in biomechanical analysis \cite{layermodelexample,geldof2022layer}.
Under these assumptions, the stiffness operator $\mathcal{K}$ in \cref{eq:harmonic_elastic_wave_equation} only depends on the elastic modulus $E$ (a.k.a. stiffness) of the soft tissue, and the spatial domain $\Omega$ on which we solve \cref{eq:harmonic_elastic_wave_equation} only depends on the thickness $T$.
This mean we can fully determine the dispersion relation based on $T$ and $E$, so we denote the dispersion relation as
\begin{equation}
    \mathfrak{D}(T, E) = \{\omega_i(\gamma)\}_{i=1}^N\quad\forall\gamma\in[0,\pi/a],
\end{equation}
where $N$ is the number of branches, or the number of eigenvalue solutions computed for \cref{eq:harmonic_elastic_wave_equation}.

\begin{figure}
    \centering
    \includegraphics[width=\linewidth]{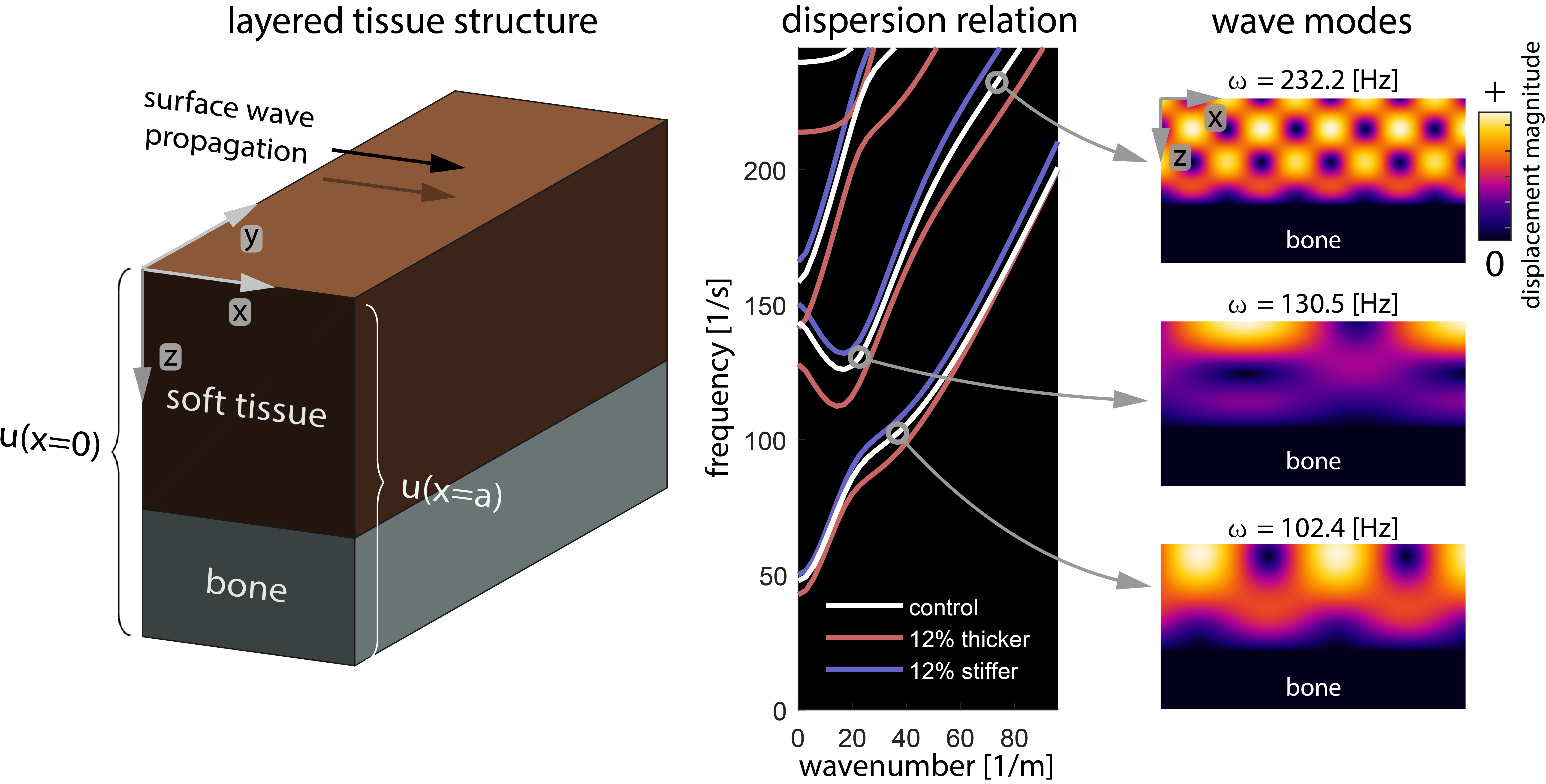}
    \caption{
    The dispersion relation depends on both the thickness and the stiffness of the soft tissue.
    We show how the dispersion relation changes when the soft tissue is made slightly thicker or slightly stiffer, with examples of the wave modes that are embedded in each dispersion relation.
    % While the wave modes look similar, subtle differences in their wavenumber and frequency content contain clues about the medium they travel in.
    Wave mode dynamics occurring beneath the surface affect their expression on the surface.
    }
    \label{fig:dispersion_relations}
\end{figure}

The dispersion relation is only concerned with the spatial ($\gamma$) and temporal ($\omega$) frequencies of the mode shapes rather than the mode shapes $\mathbf{u}(\mathbf{x})$ themselves.
% This means that for waves that express themselves, even in part, on the surface we can extract spatial and temporal frequency information from just surface motion and obtain a dispersion relation.
% This means that we can extract spatial and temporal frequency information from waves that express themselves on the surface and obtain a dispersion relation.
From waves that express themselves, even in part, on the surface, we can extract spatial and temporal frequency information to obtain a dispersion relation.
\cref{fig:dispersion_relations} shows how the dispersion relation changes with the thickness and stiffness of the soft tissue layer.
\textit{Our method leverages the sensitivity of the dispersion relation to small changes in these parameters.}

\section{Methods}
\begin{figure*}
    \centering
    \includegraphics[width=\linewidth]{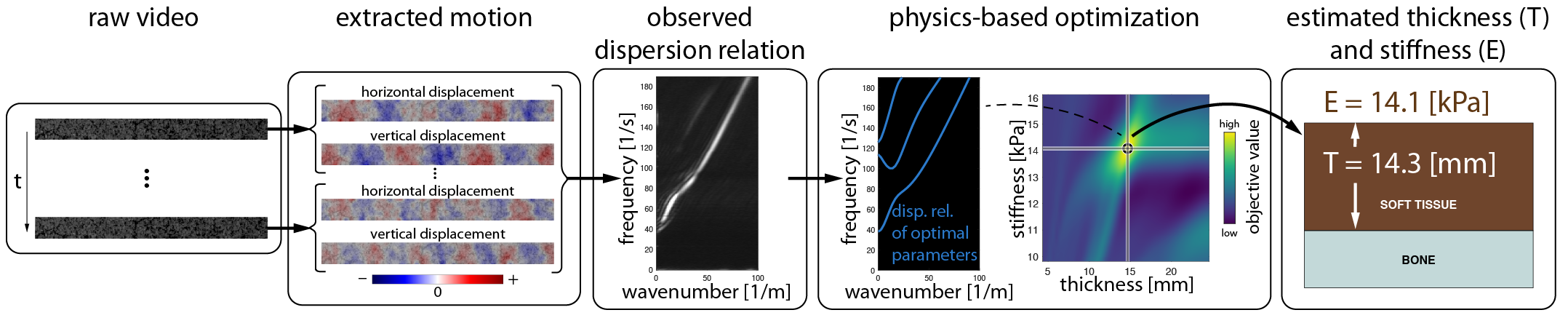}
    \caption{Method overview. 
    Given a video of the surface of the medium of interest, we first extract motion fields in image-space.
    From these we extract the observed dispersion relation, which dictates the spatial and temporal frequencies of waves that travel through the medium. 
    We solve a physics-based optimization problem to estimate the thickness and stiffness values that lead to a theoretical dispersion relation that best agrees with the observed dispersion relation.
    }
    \label{fig:method}
\end{figure*}

This section describes the computational method of VSWE.
The input is a video of motion on the surface of the medium of interest, and the output is the estimated thickness and stiffness of the medium.
In between there are two broad stages: (1) extracting a dispersion relation from the video and (2) solving for the tissue properties that best agree with the observed dispersion relation.
\cref{fig:method} presents an overview of the pipeline.
% We discuss practical considerations, such as how to excite surface waves and obtain a video, in \cref{sec:experiments}.
\textbf{Check out the supplemental material for an interactive demo of how the method works.}

% \subsection{Capturing surface waves with a video camera}
% In order to observe surface waves, we first need to provide a force to the medium of interest.
% In theory we can cause surface waves with various excitations, such as quickly tapping the surface. \berthy{better examples?}
% In our work, we use a shaker to apply a vertical excitation on the surface.
% We excite waves over a range of frequencies via a \textit{chirp} excitation, which is a sinusoidal signal that has a continuously time-varying frequency.
% This encourages a more comprehensive expression of the dispersion relation by covering a wide range of temporal frequencies.
% \berthy{discuss passive forces?}

% In practice, to ensure that the system does not experiences impulses (which may cause unexpected behavior both numerically and experimentally), we make sure to turn the chirp on and off gradually.
% We use a smoothed heaviside function with a continuous second derivative to achieve this (\TODO{move to / include details in appendix}.
\subsection{Extracting a dispersion relation from video}
\subsubsection{Motion extraction}
The first step is to quantify the image-space displacements in the video.
Since surface wave motion tends to be small, we use phase-based motion processing \cite{wadhwa2013phase}, which is sensitive to sub-pixel displacements.
Specifically, we compute local phase shifts in a complex steerable pyramid \cite{simoncelli1995steerable} and then convert these to pixel displacements \cite{fleet1990computation}.
The result is the image-space horizontal displacement $\tilde{u}(\tilde{x},\tilde{y},t)$ and vertical displacement $\tilde{v}(\tilde{x},\tilde{y},t)$ at each pixel $(\tilde{x},\tilde{y})$ and video frame $t$, where the displacements are relative to the first frame.
(Image-space coordinates are notated with a tilde to avoid confusion with world-space coordinates.)

\subsubsection{Estimating the dispersion relation}
\begin{figure*}
    \centering
    \includegraphics[width=0.9\linewidth]{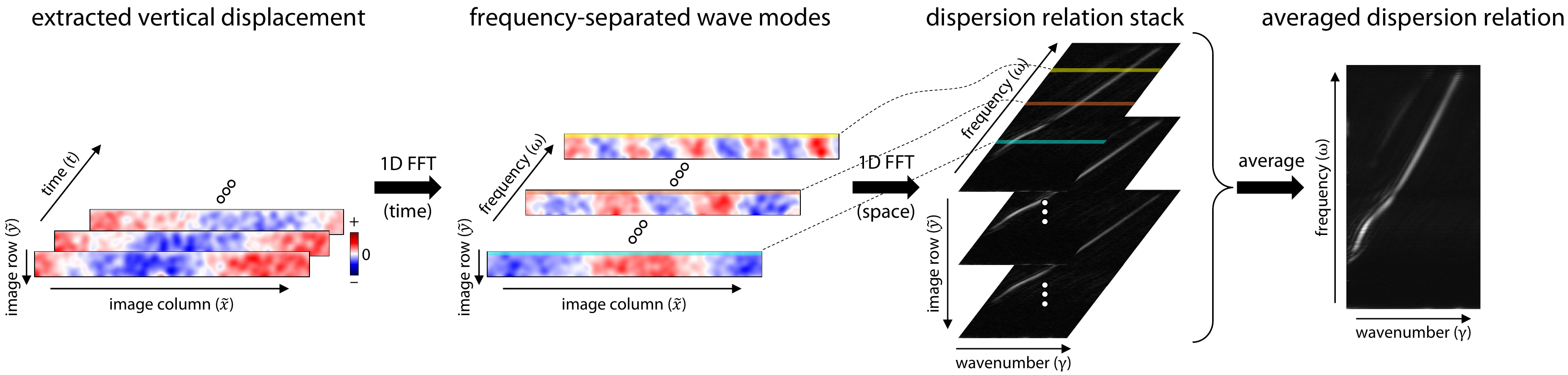}
    \caption{Obtaining a dispersion relation from image-space motion.
    Here we demonstrate with the vertical displacements taken from a real video of a gelatin sample.
    Taking a 1D FFT across time for every pixel allows us to separate wave modes by frequency ($\omega$).
    Taking another 1D FFT across space (in the $\tilde{x}$ direction) decomposes the wave modes into spatial frequencies ($\gamma$).
    The result is a 2D FFT representing the dispersion relation.
    We average the dispersion relations across image rows and both displacement directions ($\tilde{u}$ and $\tilde{v}$).
    % We note that the displacements here look similar to the frequency-separated wave modes because we used a chirp signal to excite the sample, so frequencies are already somewhat separated in time.
    % However, in general, a video may contain a mix of frequencies at any given time.
    }
    \label{fig:fft}
\end{figure*}

We process the image-space surface displacements into a dispersion relation via the fast Fourier transform (FFT).
We assume that the waves are traveling in the horizontal direction in image-space.
For each row of pixels in the $\tilde{u}(\tilde{x},\tilde{y},t)$ video, we take a 2D FFT, transforming the space dimension $\tilde{x}$ to the wavenumber dimension $\gamma$ and the time dimension $t$ to the frequency dimension $\omega$.
That is, for the row where $\tilde{y}=\tilde{y}_i$, we convert the $\tilde{u}(\tilde{x},\tilde{y}=\tilde{y}_i,t)$ signal to the complex-valued signal $\widehat{\tilde{u}}(i)(\gamma,\omega)$, whose magnitude gives the dispersion relation.
To improve the signal-to-noise ratio, we average across all the rows and both displacement directions, resulting in the observed dispersion relation $\mathbf{D}_\text{obs}$:
% \begin{equation}
%     \mathbf{D}_\text{obs}:=\left\lVert \frac{1}{2H}\sum_{i=1}^H\left(\widehat{\tilde{u}}(i)(\gamma,\omega)+\widehat{\tilde{v}}(i)(\gamma,\omega)\right) \right\rVert,
% \end{equation}
\begin{equation}
    \mathbf{D}_\text{obs}:= \frac{1}{2H}\sum_{i=1}^H\left(\left\lvert\widehat{\tilde{u}}(i)\right\rvert+\left\lvert\widehat{\tilde{v}}(i)\right\rvert\right) ,
\end{equation}
where $H$ is the number of rows in the image.
\cref{fig:fft} illustrates this process, showing how the 2D FFT pulls out spatial and temporal frequency content.

The main challenge is that the observed dispersion relation likely does not fully agree with the true theoretical dispersion relation of the medium.
First of all, it is incomplete: wave modes that do not prominently express on the surface or require large amounts of energy to propagate may not appear strongly in the video, and frequencies above the Nyquist sampling rate of the video cannot be captured.
Second of all, the extracted motion may be noisy due to camera noise or spurious motion (from, e.g., lights flickering or rigid-body motion).
VSWE has some margin of error due to the discrepancy between the observed and true dispersion relations, although we show in our experiments in \cref{subsec:planestrain} it is sensitive to $5\%$ changes in the true parameters.

% \begin{figure}
%     \centering
%     \includegraphics[width=\linewidth]{fig/fft_diagram.png}
%     \caption{Visualization of obtaining a dispersion relation from motion extracted from a video. Here $\tilde{u}_t$ is the image of horizontal displacements at frame $t$, relative to the first frame. For every row in the image, we take a 2D FFT across the width of the image and across all frames to obtain a dispersion relation. We average across the dispersion relations obtained from all the rows. We do this process with the vertical displacements $\tilde{v}_t$ and average the two resulting averaged dispersion relations.
%     }
%     \label{fig:fft_diagram}
% \end{figure}

\subsection{Estimating the thickness and stiffness}
\label{subsec:inference}
The next stage is to find the geometric and mechanical parameters that best explain the observed dispersion relation.
We do this via physical simulation and optimization.
For a given hypothesized thickness $T$ and stiffness $E$, we employ the finite element method (FEM) to numerically compute the dispersion relation.
We wrote specialized FEM code to do so efficiently and will release the code upon publication.
% \footnote{\url{https://github.com/aco8ogren/tissue-dispersion}}
% \alex{include benchmark comparison against comsol? to quell reviewer fears about ``solving PDE takes a long time?"} \katie{put benchmark comparison in footnote if necessary}.
The goal is to maximize an objective function that rewards similarity between $\mathfrak{D}(T,E)$ and $\mathbf{D}_\text{obs}$.

After testing various objective functions, we found that treating dispersion relations as images and using the structural similarity index measure (SSIM) \cite{wangImageQualityAssessment2004} to work best.
That is, we aim to maximize the SSIM between the images of the observed and proposed dispersion relations.
The observed dispersion relation $\mathbf{D}_\text{obs}$, since it is derived via a 2D FFT, is already represented as an image.
However, the physics-based dispersion relation $\mathfrak{D}(T,E)$ is computed as a set of curves.
We transform $\mathfrak{D}(T,E)$ into an image $\mathbf{D}_\text{hyp}(T,E)$ by assigning intensities with a Gaussian kernel based on the distance from each $(\gamma,\omega)$ pixel to the curves.
Then we solve the following optimization problem:
\begin{equation}
\label{eq:optimization_problem}
    T^*,E^*=\argmax_{T,E}\text{SSIM}\left(\mathbf{D}_\text{hyp}(T,E),\mathbf{D}_\text{obs}\right).
\end{equation}
We solve \cref{eq:optimization_problem} with a grid search over possible thickness and stiffness values, although more efficient approaches can be taken for computationally demanding settings.

\section{Results}
\label{sec:experiments}
% Sensitivity analysis with plane strain
% Real data with plane strain
% 3D anatomical
% Loss function ablation
% Experimental considerations
We validate our approach on real and simulated data. We demonstrate remarkable accuracy on real gelatin samples, and we demonstrate recovering spatially-varying thickness and stiffness across a simulated 3D human leg.

\subsection{Simulated plane strain}
\label{subsec:planestrain}
We created a two-layer tissue model in COMSOL \cite{comsol} that would allow us to simulate samples with ground-truth geometric and mechanical parameters.
We set stiffness values on the order of $10$ \si{kPa}, which is a reasonable range for soft tissue \cite{egorov2008soft}.
We simulated the model's response to a chirp excitation signal applied to the leftmost side of the surface of the sample.
The simulations were done with the plane-strain assumption, which allows for modeling 3D dynamics with only two dimensions by assuming that strain is constant in the $z$ direction.
We used the simulated horizontal and vertical displacements to obtain dispersion relations.

\begin{figure}
    \centering
    \includegraphics[width=0.6\linewidth]{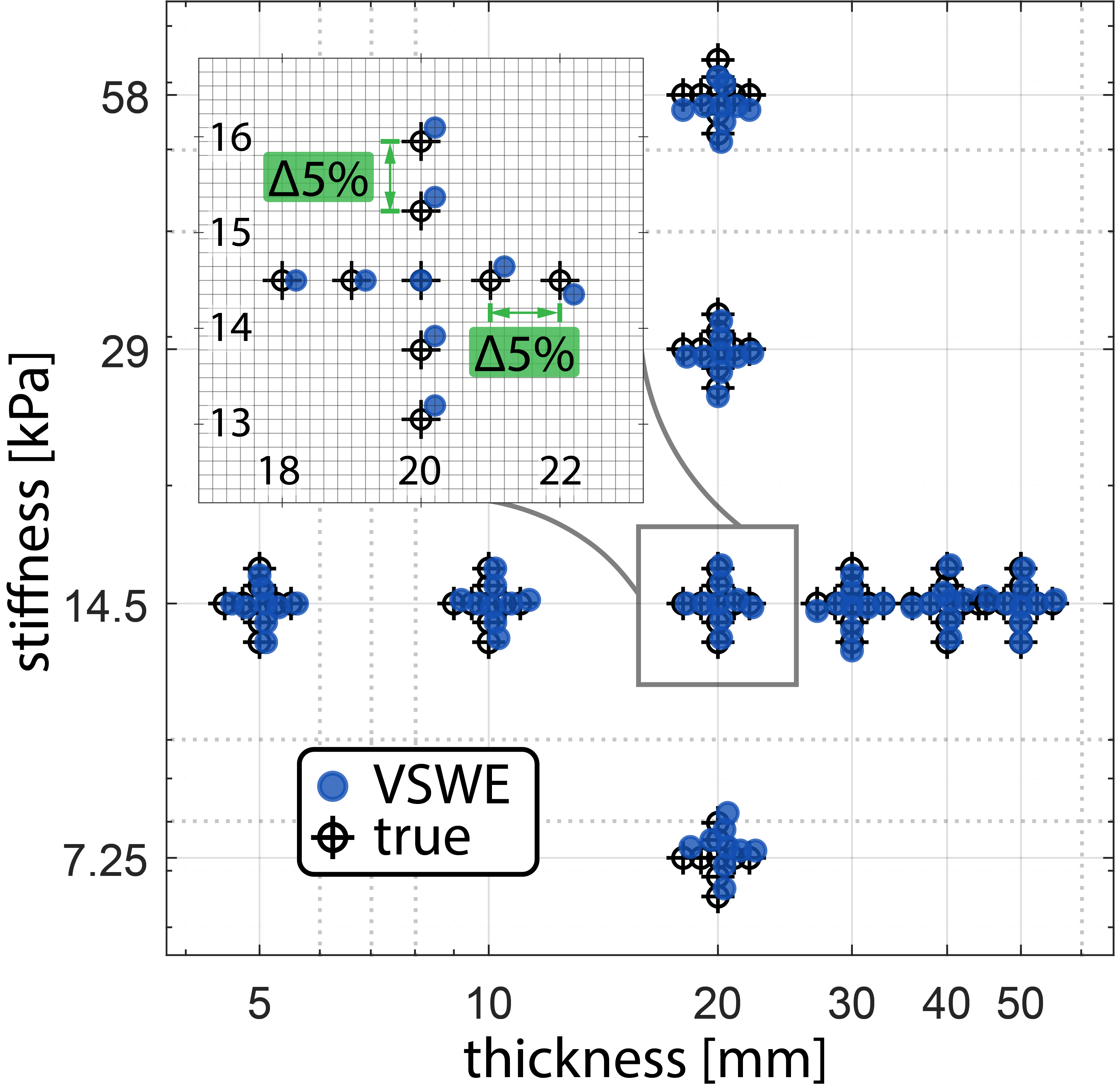}
    \caption{Sensitivity analysis.
    We applied VSWE to plane-strain simulations with a range of true thickness and stiffness values, along with slight perturbations of those values.
    In each cluster, we perturbed the true parameters by $\pm5\%$ and $\pm 10\%$ from the central values and found that VSWE was sensitive to these changes.
    }
    \label{fig:plane_strain_sensitivity}
\end{figure}

We tested the sensitivity of our method by applying slight changes to the true thickness and stiffness of the simulated sample.
That is, for a certain thickness and stiffness, we can simulate samples assuming those parameters as well as perturbations of those parameters and assess whether VSWE picks up on these slight changes.
As an experiment, we simulated samples at 5 distinct thicknesses and 4 distinct moduli.
For each thickness-stiffness pair, we looked at $5\%$ and $10\%$ perturbations of either parameter. This led to 9 clusters with 9 simulated samples each. 
As \cref{fig:plane_strain_sensitivity} shows, within each cluster, the estimated parameters track with changes in the true parameters, even when the true parameters have changed by only $5\%$.

\subsection{Real videos of gelatin-based phantoms}
To test our method on real data, we created gelatin-based phantoms (i.e., samples mimicking biological tissue) of varying thicknesses.
We poured varying amounts of gelatin ($1000$, $1100$, and $1500$ \si{mL}) into the same-sized container, leading to three different thicknesses.
We set the samples in the refrigerator for about 24 hours.
Once they were set, we sprinkled garlic powder onto the samples to create texture for motion extraction.

\begin{figure}
    \centering
    \includegraphics[width=\linewidth]{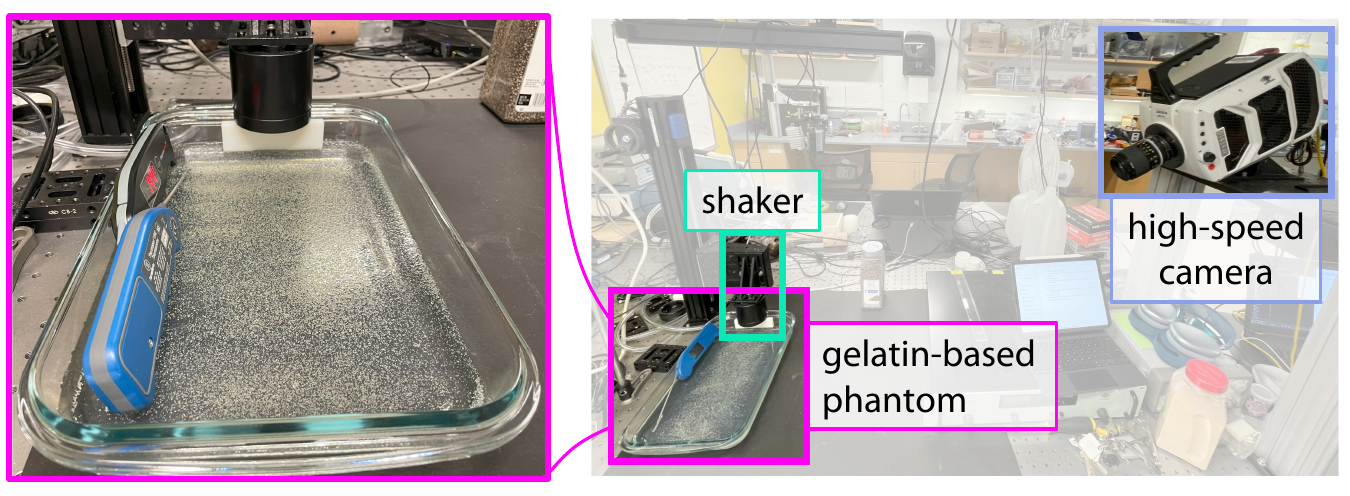}
    \caption{Experiment setup. A shaker was applied on one side of the gelatin-based phantom to excite waves in the medium. A high-speed camera, zoomed in on the phantom's surface, captured videos of the response to the shaker. As the left picture shows, we took temperature recordings with a thermometer.}
    \label{fig:experiment_setup}
\end{figure}

For each set sample, we measured the thickness with calipers, and we used rheometry to obtain ground-truth stiffness values.
To obtain videos, we applied a shaker at one end of the sample to excite waves with a chirp signal and recorded the surface of the sample with a high-speed camera at 600 FPS (each video was about four seconds long).
\cref{fig:experiment_setup} shows pictures of the experiment setup.
% Fig.~\ref{fig:real_data} shows several frames of one of the videos, along with the corresponding extracted horizontal and vertical displacements.

% \begin{figure*}
%     \centering
%     \includegraphics[width=0.8\linewidth]{fig/motion_extraction.png}
%     \caption{Real data. \TODO{add x-t diagram and dispersion relation visual. Swap out the motion for a sample with a more 'photogenic' dispersion relation?}}
%     \label{fig:real_data}
% \end{figure*}

\begin{figure}
    \centering
    \includegraphics[width=\linewidth]{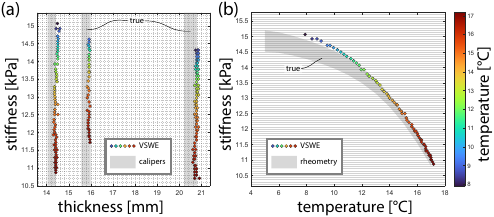}
    \caption{
    \textbf{(a) Estimated thickness and stiffness for three phantoms over a range of temperatures.}
    The phantoms were produced with three different volumes ($1000$, $1100$, $1500$ \si{mL}) of gelatin but set in the same-sized container, thus generating three different thicknesses.
    The thickness of each sample was measured with calipers.
    We produced a confidence interval of the ground-truth thickness by taking the $0.2$ and $0.8$ quantiles of multiple calipers measurements.
    The VSWE-estimated thickness consistently falls within the confidence interval.
    Note that while each phantom was polymerized according to the same approximate recipe, slight variations in the recipe, along with slightly different polymerization durations, led to some variation in the stiffness across samples.
    The temperature changes occurred as the samples spent more time at room temperature after being removed from the fridge. 
    \textbf{(b) Estimated stiffness values for the thinnest sample over the same range of temperatures.}
    The ground-truth elastic modulus was measured with rheometry over a similar temperature range.
    Because the modulus of hydrogels can depend on both temperature and frequency \cite{zorner2010measurement}, the rheometry measurements show a different stiffness range at each temperature.
    Note that the rheometry measurements were only taken between $10$--$100$ \si{Hz} due to instrument limitations, while the excitation signal ranged from $40$--$200$ \si{Hz}.
    This may explain some of the discrepancy between VSWE and rheometry, especially for the colder (and thus stiffer) samples which exhibit a stronger expression of higher-frequency wave dynamics.
    Regardless, in each case, VSWE estimates the stiffness extremely well, within $1.2\%$ error of the rheometry range.
    }
    \label{fig:rheometry_comparison}
\end{figure}

For each of the three samples, we took many rheometry measurements and videos over the course of about one hour after removing the sample from the refrigerator.
In total we obtained about 60 videos per sample over time, each with corresponding ground-truth parameters.
\cref{fig:rheometry_comparison} shows the inferred parameters for each sample and each point in time.
\cref{fig:rheometry_comparison}(a) shows that our method clearly identifies the three different thicknesses of the samples. As a sample spends more time out of the refrigerator, its temperature increases and its stiffness decreases. \cref{fig:rheometry_comparison}(b) shows that the estimated stiffness decreases accordingly with temperature.
% We note that the estimated stiffness is sometimes higher than the range obtained with rheometry because 

\subsection{3D human leg with spatially-varying thickness}
\label{sec:anatomical}
As another step towards realism, we tested VSWE on a simulated female human leg.
The anatomical geometry of the leg was obtained as an STL file from the dataset of \citet{andreassen2023three}, who created 3D models from the National Library of Medicine's Visible Human Project \cite{visiblehumanproject}.
We simulated the leg's response to a chirp excitation applied on the leg in COMSOL, running a full 3D physics simulation without any assumptions besides linear elasticity.
For the sake of computational feasibility, we ran the simulation on the lower half of the leg, which we refer to as the calf region.
To obtain dispersion relations, we considered the simulated displacements in two directions: one tangent to the leg's surface and one normal to the leg's surface.

\begin{figure*}
    \centering
    \includegraphics[width=\linewidth]{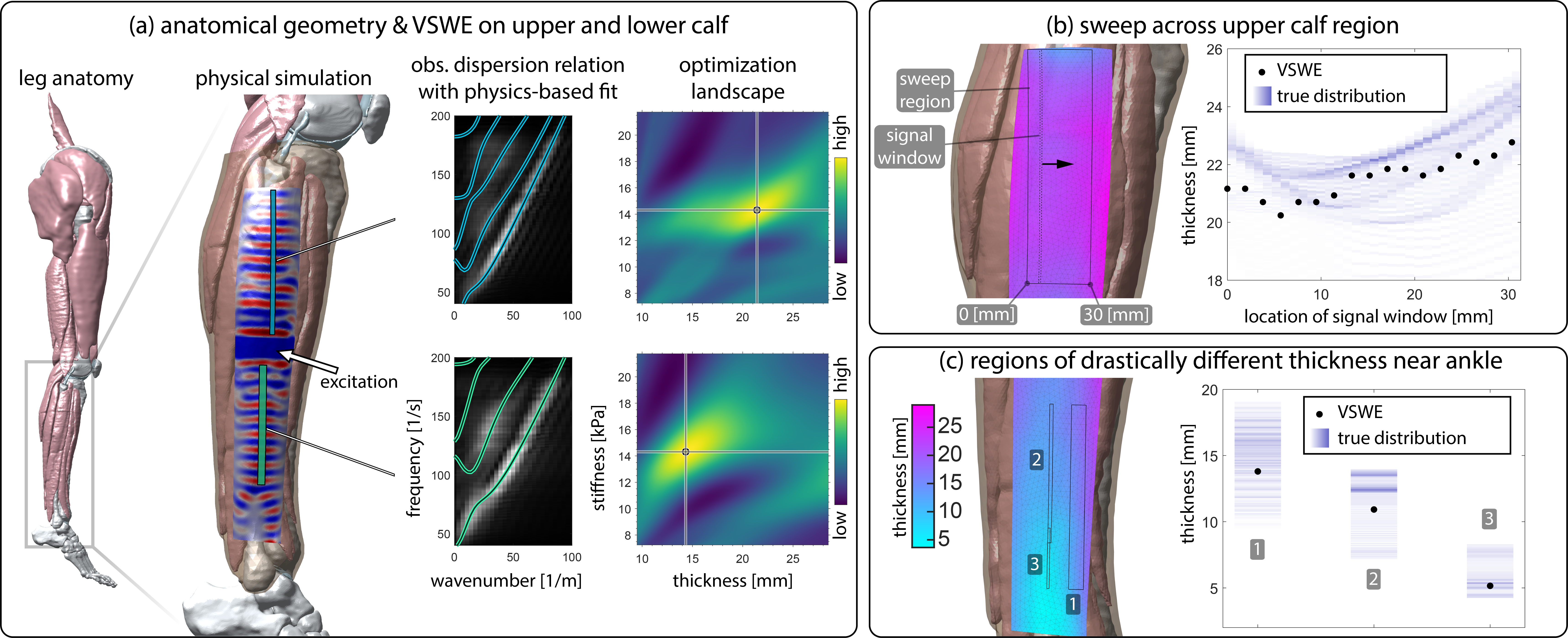}
    \caption{3D anatomical inference. Panel
    \textbf{(a)} shows the realistic 3D leg anatomy (taken from the Visible Human Project \cite{visiblehumanproject}). We ran a full 3D simulation of the leg's response to a chirp excitation applied on the skin. The observed dispersion relation, along with the optimization landscape and hypothesized dispersion relation of the optimal parameters, is shown for two regions: one on the upper calf and one on the lower calf near the ankle. Panel \textbf{(b)} shows results of sweeping an observation window across a region of the upper calf. A ground-truth thickness distribution was obtained by computing the distance from the skin to the nearest point on the bone across the observation window. The VSWE-estimated thickness as the observation window slides from left to right tracks with the changing thickness of the leg. Panel \textbf{(c)} shows results for three distinct regions near the ankle. The VSWE estimations reflect the drastically different thicknesses of these regions.
    }
    \label{fig:anatomical}
\end{figure*}

We applied a sweeping window across the upper calf region, estimating the thickness and modulus at each window location.
% in the direction along the circumference of the bone (i.e., the window spans the entire length of the calf but has a small width).
\cref{fig:anatomical} shows how, as the window sweeps around the leg, the inferred thickness changes. 
We computed the true thickness by taking the distance from the point on the skin to the nearest point on the bone. 
This results in a thickness distribution within the window, since the thickness changes slightly in the lengthwise direction of the leg, as well.
\cref{fig:anatomical} shows the inferred thickness and the distribution of ground-truth thicknesses for each location of the sweeping window.
We find that the estimated thickness roughly agrees with the true thickness distribution.
There appears to be slightly more error near the edges of the simulated domain, which may be due to boundary effects.
While the physics-based simulation in COMSOL is well-modeled in the interior of the domain, there may be artifacts near the boundary of the domain, making it harder to infer the correct tissue parameters near the boundary.

We also applied VSWE to three different windows in the lower calf region.
Rather than perform a sweep in this region, we chose to focus on three distinct windows since there are three subregions in the lower calf area that have notably different thicknesses, due to the tissue structure changing a lot near the ankle.
As \cref{fig:anatomical} shows, the inferred thickness agrees with the true thickness distribution for each subregion.
Our method also recovered the constant stiffness throughout the calf, as shown in \cref{fig:teaser}.

\subsubsection{Objective function ablation}
\begin{figure}
    \centering
    \includegraphics[width=\linewidth]{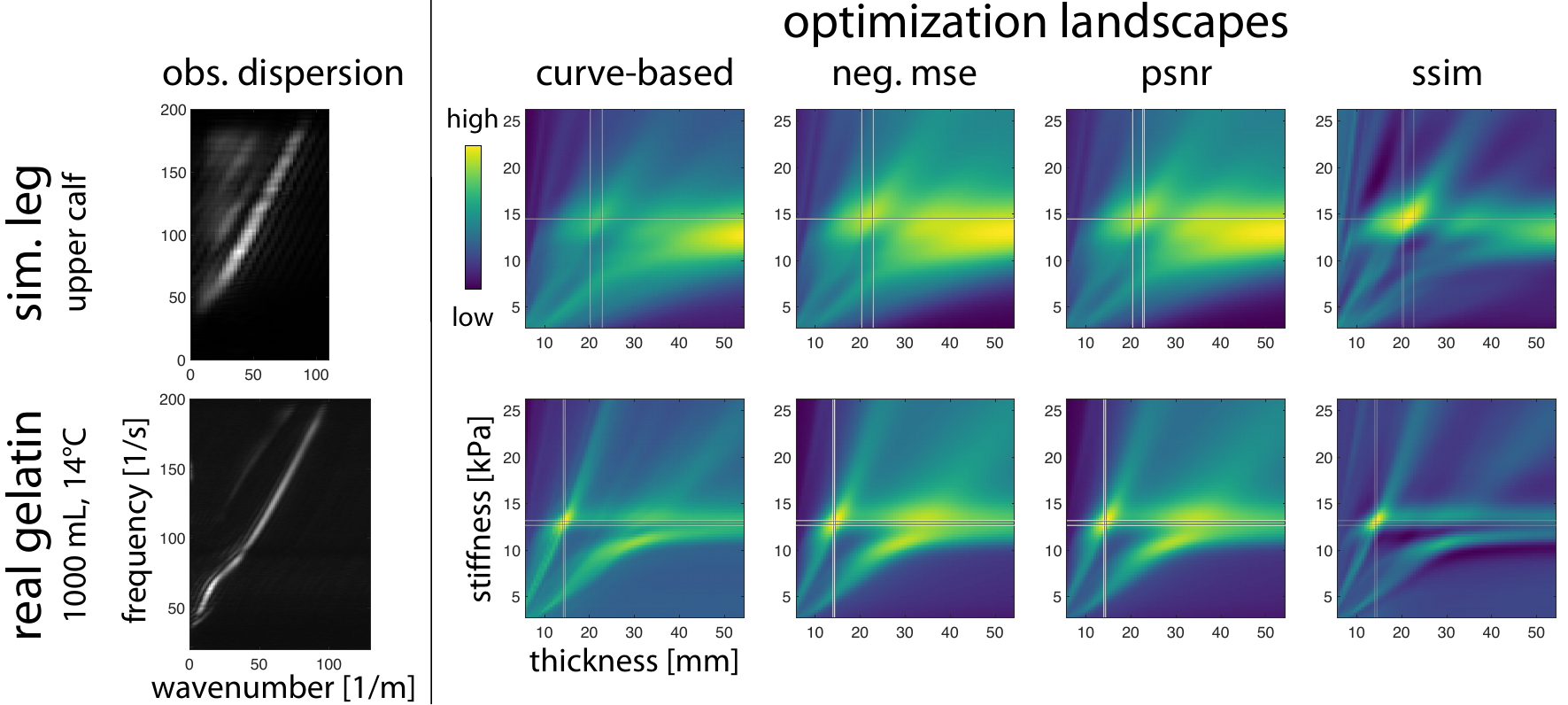}
    \caption{
    Ablation of optimization objective function.
    The optimization landscape is shown for various objective functions, applied in two different scenarios.
    % the simulated upper calf (top row) and the thinnest real gelatin sample (bottom row).
    Each landscape is visualized with its own color map limits, but bright yellow and dark blue indicate high (good) and low (bad) objective values, respectively.
    SSIM gives the sharpest optimization landscape.
    For the simulated upper calf, all the objective functions besides SSIM lead to the wrong optimal parameters (cross hairs indicate target parameters).
    Additionally, note how the simulated upper calf leads to a blurrier dispersion relation and hence blurrier optimization landscapes than the real gelatin sample, which has a simple geometry.
    }
    \label{fig:loss_ablation}
\end{figure}

As mentioned in \cref{subsec:inference}, we find that SSIM performs best for the optimization objective function.
This choice was crucial for obtaining acceptable results with the 3D leg.
\cref{fig:loss_ablation} shows the optimization landscape for different choices of the objective function, demonstrated on both the upper calf simulation and a real video of gelatin.
We compare SSIM to the following objective functions: a curve-based objective function, the negative mean squared error (MSE) between the images, and the peak signal-to-noise ratio (PSNR) between the images.
The curve-based loss function goes through all the curves (i.e., $\{\omega_i(\gamma)\}_{i=1}^N\forall\gamma\in[0,\pi/a]$) in $\mathfrak{D}(T,E)$ and integrates the values of the observed dispersion relation $\mathbf{D}_\text{obs}$ along the curves.
\cref{fig:loss_ablation} shows that SSIM leads to the sharpest loss landscape and the most accurate estimated parameters.

\section{Discussion}
\begin{figure}
    \centering
    \includegraphics[width=0.9\linewidth]{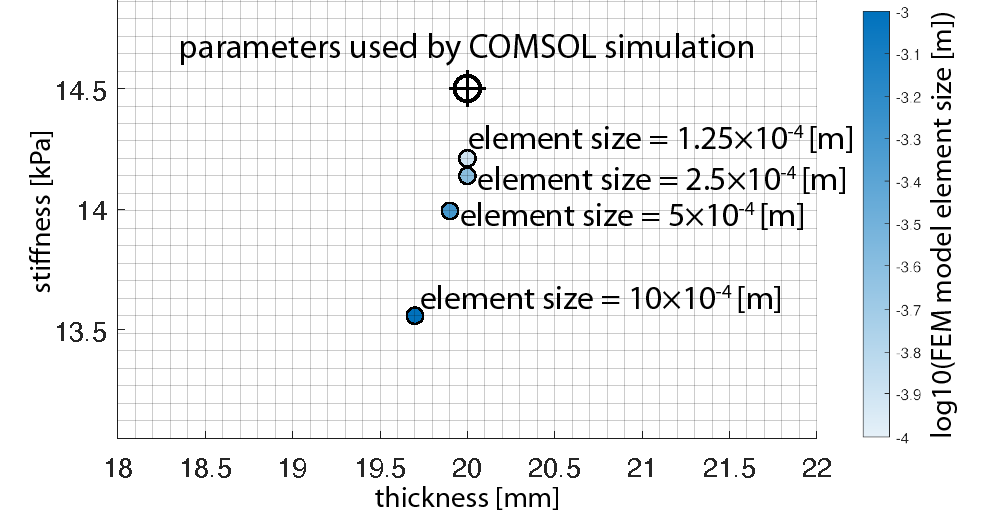}
    \caption{Ablation of FEM mesh resolution.
    As the mesh becomes finer, the VSWE-estimated parameters approach the true parameters used for the COMSOL plane-strain simulation.
    }
    \label{fig:mesh_mismatch_ablation}
\end{figure}

\begin{figure}
    \centering
    \includegraphics[width=\linewidth]{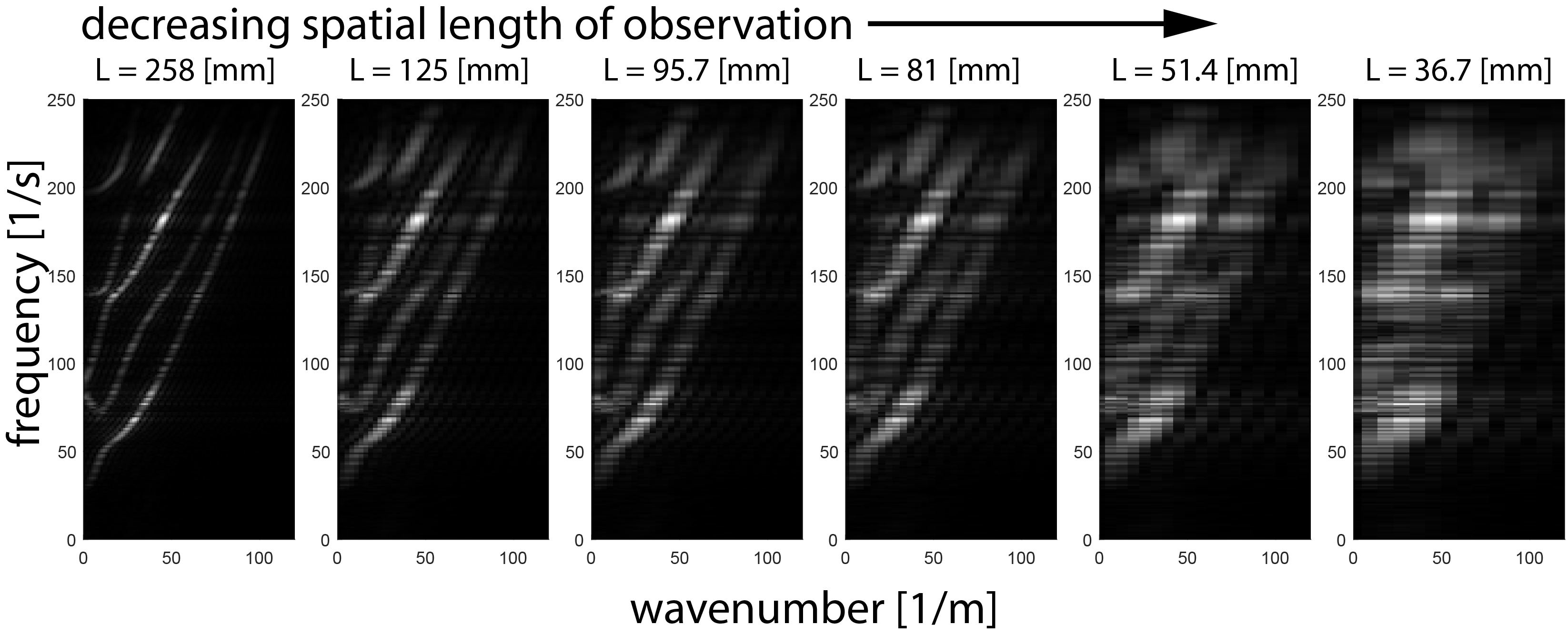}
    \caption{
    Ablation of spatial extent of observation.
    Keeping all else fixed, decreasing the spatial length $L$ of the observation window degrades the quality of the observed dispersion relation. A degraded dispersion relation in turn leads to a fuzzy optimization landscape (see \cref{fig:loss_ablation} for an example).
    }
    \label{fig:domsize_ablation}
\end{figure}
\subsection{Practical considerations}
VSWE's performance depends on many factors, including the types of waves supported by the tissue, the quality of the observation, and the quality of the physics model.
One consideration is the resolution of the FEM mesh used to model physics, as the element size determines the modeling accuracy and the smallest modeled wavelength.
\cref{fig:mesh_mismatch_ablation} shows that as the element size decreases, the estimated parameters approach the true parameters.
% The consequence is that the estimated parameters may be less accurate.
% (see Fig.~\ref{fig:loss_ablation} for a visualization of how a blurrier observed dispersion relation leads to a blurrier optimization landscape).

Another consideration is that the spatial length $L$ of the observation window determines the largest wavelength and the number of wavelengths that can be observed.
\cref{fig:domsize_ablation} shows how, keeping all else fixed, the FFT degrades as the observed domain shrinks, which makes inference more challenging.

\subsection{Characteristic numbers}
\begin{figure}
    \centering
    \includegraphics[width=0.9\linewidth]{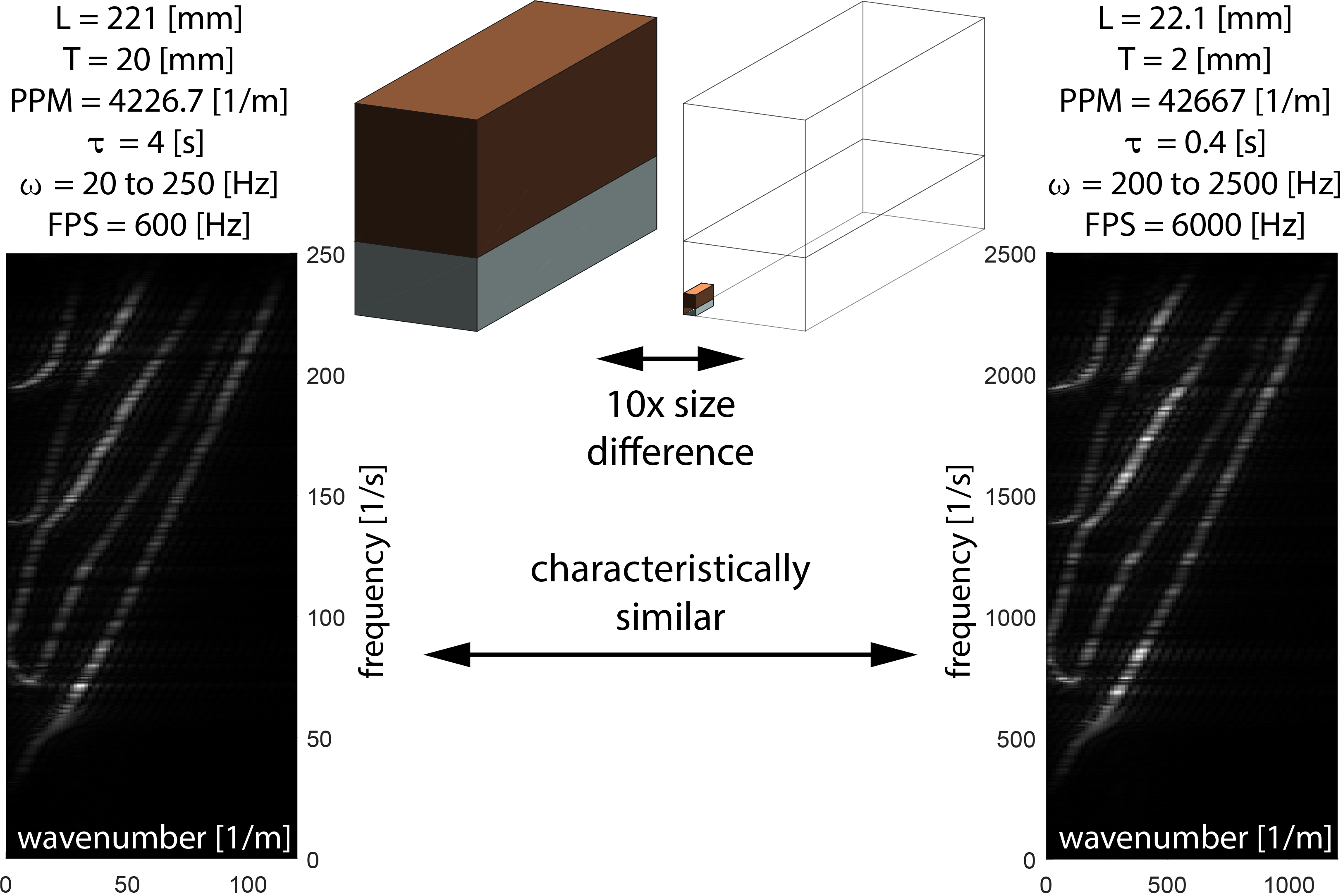}
    \caption{
    Leveraging similitude of characteristic numbers enables the application of VSWE to extremely different parameter ranges.
    In this simulated example, we show that by preserving the values of $\pi_{1-6}$, we see a similar observed signal from a system with parameters $10\times$ different than the primary ranges studied in this paper.
    In particular, we rescaled the size of the observation window, the thickness of the tissue, the spatial sampling rate, the total observation time, the frequency range of the excitation, and the temporal sampling rate.
    % Specifically, if we decrease the size of the observed domain by a factor of $10$, we can observe \textit{characteristically similar} wave dynamics by increasing the frequency range of excitation to preserve the characteristic numbers.
    % On the left, we show observation of a longer domain with thicker soft tissue at a lower frequency range.
    % On the right, we show that when we decrease the length of the domain and thickness of the soft tissue, we can recover the same effective dynamics by increasing the frequency range proportionally.
    Note that the observation window on the right is smaller than that of the most degraded FFT in \cref{fig:domsize_ablation}, yet the FFT here has not degraded.
    % In short, even though we decreased L, we kept $\pi_1$ constant by decreasing $\gamma$ (through increasing $\omega = c(E,\rho)/\omega$) which preserved our spatial FFT resolution, and kept $\pi_4 = \gamma T$ constant by reducing $T$ which preserved the wave behavior, ultimately resulting in the same dispersion signal (in both quality and physical behavior), but at a vastly different scale.
    }
    \label{fig:nondim_scaling}
\end{figure}
In our ablation studies in \cref{fig:domsize_ablation} and \cref{fig:mesh_mismatch_ablation}, we kept all other parameters of the problem fixed.
By changing other parameters (e.g., increasing the image resolution, changing the range of frequencies in the excitation signal, or changing the tissue thickness), we can overcome certain limitations.
This leads us to introduce \textit{characteristic numbers}.

We can characterize the physical system and inference problem with characteristic numbers (a.k.a. dimensionless numbers or $\pi-$groups).
Some characteristic numbers reflect the observability of wave modes, such as $\pi_1:=\gamma L$, $\pi_2:=\text{PPM}/\gamma$, $\pi_3:=\omega\tau$, and $\pi_4:=\text{FPS}/\omega$, where PPM (pixels per meter) is the spatial sampling rate, FPS (frames per second) is the temporal sampling rate, and $\tau$ is the total observation time.
Increasing $\pi_{1-4}$ improves VSWE's performance by improving the quality of the FFT dispersion.\footnote{Remark: $\omega$ and $\gamma$ are roughly coupled through the wavespeed $c \sim \sqrt{E/\rho} \sim \omega/\gamma$ of the tissue, meaning that $\pi_{1-2}$ and $\pi_{3-4}$ are related via the material properties.}

Other characteristic numbers reflect the numerical accuracy of the FEM model used to fit the observation, such as $\pi_5:=\frac{1}{\gamma e}$, where $e$ is the FEM element size.
% $\pi_6:=\frac{1}{\omega\delta t}$
Increasing $\pi_5$ improves performance by ensuring that physics is well-modeled in the parameter-fitting process.

Other characteristic numbers reflect the actual physics that are possible in the system.
For example, $\pi_6:=\gamma T$ conveys the ``shallowness'' of the wave, measuring how significantly the stiff foundation affects the dynamics of a surface wave.
It is beneficial for $\pi_6$ to be neither too large nor too small.
If $\pi_6$ is too large, then waves behave similarly to how they would in an infinitely-thick medium, making it difficult to pinpoint the thickness.
% and all observed wave solutions will be effectively equivalent to the Rayleigh wave half-plane solution.
If $\pi_6$ is too small, waves above a certain wavelength will not be permitted to exist, making it necessary to capture smaller wavelengths.\footnote{A small $\pi_6$ explains why ocean waves break near the shore. It is also the reason that a higher-frequency excitation was helpful for the very thin tissue near the ankle in \cref{fig:anatomical}.}

% \alex{positive spin? katie did say to make sure we are being positive about our characteristic numbers, and all I did was highlight bad cases in this paragraph lol}
% In between, there is a substantial sweet spot where waves exist and are notably affected by the dynamics 

Understanding these characteristic numbers is important for extending the use of VSWE to scenarios and applications with vastly different parameter ranges than those studied in this paper, whether the objects are much larger or smaller, stiffer or softer, denser or lighter, or exhibit dynamics at much higher or lower wavenumbers or frequencies.
In general, by preserving the values of these characteristic numbers (e.g., by increasing FPS in coordination with $\omega$), we can preserve the performance of VSWE.
% \footnote{While we have provided some examples of important characteristic numbers, they do not represent a comprehensive list.}

\cref{fig:nondim_scaling} provides an example of how to take advantage of characteristic numbers.
Specifically, we overcome the challenge of a smaller observed domain (as illustrated in \cref{fig:domsize_ablation}) by rescaling other parameters to preserve the relevant characteristic numbers and thus maintain the quality of the observed dispersion relation.
% (i.e., the spatial sampling rate, the thickness, the frequency range of the excitation, the total observation time, and the temporal sampling rate)
% Specifically, if we decrease the size of the observed domain by a factor of $10$, we can observe \textit{characteristically similar} wave dynamics by increasing the frequency range of excitation to preserve the characteristic numbers.
In general, one should look to characteristic numbers as guidelines for adjusting system and/or inference parameters in order to maintain the performance of VSWE across a variety of scenarios.
% Of particular interest may be manipulating other system parameters to maintain the efficacy of VSWT for smaller observation windows $L$, sample rates FPS, or quite large or small tissue thickness $T$ or stiffness $E$.

% A good rule of thumb is $\pi_1>\TODO{give concrete guidance for setting $L$, FPS, element size}

\section{Conclusion}
We have introduced Visual Surface Wave Elastography for estimating subsurface geometric and mechanical properties from visible surface waves.
Our method works by extracting spatial and temporal frequency information from surface waves observed in video.
By representing this information in the form of a dispersion relation, we can use a fully physics-based approach to solve for the properties that explain the wave behavior observed in the video.
We validated our method on both simulated and real data, showing that our method is sensitive to small changes in the properties of tissue-like media.
We demonstrated the promise of applying VSWE in the real world by testing it on real videos of gelatin-based phantoms and on a simulated realistic 3D human leg.
% We discussed practical requirements when dealing with particularly small/large domains or particularly stiff/soft tissue, including the frame rate of the camera and the frequency of excitation.
Our work shows the potential of everyday visual data to reveal critical information deep beneath the skin.

% \alex{possible future work (if we want to include it) could be introducing more layers and determinining if it's feasible to characterize multiple layers at once.}
\section*{Acknowledgments}
ACO and CD are supported by DOE award no.~DE-SC0021253 and the New Frontiers of Sound (NewFoS) NSF center.
BTF and KLB acknowledge support from the NSF GRFP, the Amazon AI4Science Partnership Discovery Grant, and the Heritage Medical Research Institute (HMRI) Fellowship.
JA and CD are supported by the HMRI and the NSF Center to Stream Healthcare in Place (C2SHIP) award no.~2052827.

{
    \small
    \bibliographystyle{ieeenat_fullname}
    \bibliography{main}
}

\newpage
\appendix

\section{Runtime analysis}
% The bulk of computational cost of VSWT lies primarily in the computation of dispersion relations to fit the observed data.
% We found it most practical to pre-compute datasets of dispersion relations for search grids of stiffness and thickness values in reasonable ranges for the characterization of soft tissue.
% We could then re-use these datasets for any number of fitting problems without having to re-do dispersion computations.
% While we expect this cost to be significantly reduced by the employment of optimization routines more efficient than grid search, we have included below some run times that exemplify the computational cost of the aforementioned dispersion relations for the use of VSWT in this paper.
The primary computational cost of VSWT stems from calculating dispersion relations to fit observed data. To address this efficiently, we precomputed dispersion relation datasets over grids of stiffness and thickness values suitable for soft tissue characterization. These datasets can be reused for multiple fitting scenarios, eliminating the need for repeated dispersion calculations. Although employing more efficient optimization methods than grid search would substantially reduce computational cost, we provide representative runtimes below to illustrate the expense associated with computing dispersion relations in this study.

In each dispersion dataset, there are several parameters that have bearing on the cost to compute the dataset.
For the computation of a dispersion dataset containing $12$ eigenvalue branches and $60$ wavenumber values, on a grid of $41$ stiffness values ($E$) $\times$ $41$ thickness values, the runtime for various numbers of elements corresponding to different element sizes $e$ are tabulated below.
Computations were performed on an Intel Xeon CPU E5-2663 v3 with 20 cores, and all computations used less than 64 GB of memory.
While we found $e = 0.5, N_{ele} = 240$ to be sufficient for reasonable performance in our settings, using smaller $e$ with higher $N_{ele}$ may be desirable for more accuracy or to model.
The compu
\begin{itemize}
    \item $e = 0.5$ \si{mm}, $N_{ele} = 240$ had runtime of $157$ \si{s}
    \item $e = 0.25$ \si{mm}, $N_{ele} = 960$ had runtime of $470$ \si{s}
    \item $e = 0.125$ \si{mm}, $N_{ele} = 3840$ had runtime of $3169$ \si{s}
\end{itemize}
\section{Optimization objective functions}
Here we define the objective functions we considered when developing our approach. Recall that $\mathbf{D}_\text{obs}$ is the observed FFT-derived dispersion relation. For a given hypothesized $(T,E)$ pair, $\mathfrak{D}(T,E)$ is the physics-based dispersion relation, and $\mathbf{D}_\text{hyp}(T,E)$ is the image version of $\mathfrak{D}(T,E)$.

\paragraph{Curve-based objective function} 
We define a simple and intuitive curve-based objective function $f$ that assigns points for the observed dispersion relation having high magnitude at the points that exist in the hypothesized physics-based dispersion relation $\mathfrak{D}(T,E)=\{\omega(\gamma)\}_{i=1}^N\forall\gamma\in[0,\pi/a]$:
\begin{equation}
    f\left(\mathfrak{D}(T,E),\mathbf{D}_\text{obs}\right)=\sum_{i=1}^N\int_0^{\pi/a}\mathbf{D}_\text{obs}\left(\gamma,\omega_i(\gamma)\right)\mathrm{d}\gamma.
\end{equation}

\paragraph{Image-based objective functions}
To convert the dispersion relations from curve format to image format, we assign pixel values with a Gaussian kernel based on the distance from the pixel to any point on any of the curves.
More precisely, the value of a pixel located at point $(\gamma_{im},\omega_{im})$ is given by
\begin{equation}
    v(\omega_{im},\gamma_{im}) = \operatorname{exp}(\frac{-d_{min}^2}{2\sigma^2})
\end{equation}
where
\begin{equation}
    d_{min}(\omega_{im},\gamma_{im}) = \min_{\gamma_{c},i}(\sqrt{(\omega_{im} - \omega_{c}^i(\gamma_{c}))^2 + (\gamma_{im} - \gamma_{c})^2})
\end{equation}
where $\omega_{c}^i(\gamma_{c})$ denotes the dispersion relation in curve form on any band $i$, and $\sigma$ is a parameter that controls the width of the curve once converted to image form.

We use standard definitions of MSE, PSNR, and SSIM \cite{wangImageQualityAssessment2004}.

\end{document}